\documentclass{article}

\usepackage[preprint]{neurips_2023}

\usepackage[export]{adjustbox}
\usepackage[utf8]{inputenc} 
\usepackage[T1]{fontenc}    
\usepackage{url}            
\usepackage{booktabs}       
\usepackage{amsfonts}       
\usepackage{nicefrac}       
\usepackage{microtype}      
\usepackage{xcolor}         
\usepackage{graphicx}       
\usepackage{subcaption}
\usepackage{wrapfig}
\usepackage{enumitem}

\definecolor{citecolor}{HTML}{2980b9}
\definecolor{linkcolor}{HTML}{c0392b}
\usepackage[pagebackref=true,breaklinks=true,colorlinks,bookmarks=false,citecolor=citecolor,linkcolor=linkcolor]{hyperref}

\newcommand{\anon}[1]{}

\title{Towards Open-World Mobile Manipulation in Homes: Lessons from the Neurips 2023 HomeRobot Open Vocabulary Mobile Manipulation Challenge}

\author{
   \begin{tabular}{lll}
   Sriram Yenamandra$^{*2}$ & Arun Ramachandran$^{*2}$ & Mukul Khanna$^{*2}$ \\ 
   Karmesh Yadav$^{*1}$ & Jay Vakil$^{*1}$ & Andrew Melnik$^{5}$ \\
   Michael Büttner$^{5}$ & Leon Harz$^{5}$ & Lyon Brown$^{5}$ \\
   Gora Chand Nandi$^8$ & Arjun PS$^9$ & Gaurav Kumar Yadav$^8$\\
   Rahul Kala$^{10}$ & Robert Haschke$^{5}$ & Yang Luo$^{11}$ \\
   Jinxin Zhu$^{11}$ & Yansen Han$^{11}$ & Bingyi Lu$^{11}$ \\
   Xuan Gu$^{11}$ & Qinyuan Liu$^{11}$ & Yaping Zhao$^{11}$ \\
   Qiting Ye$^{11}$ & Chenxiao Dou$^{11}$ & Yansong Chua$^{11}$ \\
   Volodymyr Kuzma$^6$ & Vladyslav Humennyy$^6$ & Ruslan Partsey$^6$ \\
   Jonathan Francis$^{4,7}$ & Devendra Singh Chaplot$^{1}$ & Gunjan Chhablani$^{2}$ \\
   Alexander Clegg$^{1}$ & Theophile Gervet$^{1,4}$ & Vidhi Jain$^{4}$ \\
   Ram Ramrakhya$^{2}$ & Andrew Szot$^{2}$ & Austin Wang$^{1}$ \\
   Tsung-Yen Yang$^{1}$ & Aaron Edsinger$^{3}$ & Charlie Kemp$^{2,3}$ \\
   Binit Shah$^{3}$ & Zsolt Kira$^{2}$ & Dhruv Batra$^{1,2}$ \\
   Roozbeh Mottaghi$^{1}$ & Yonatan Bisk$^{1,4}$ & Chris Paxton$^{1}$ \\[3pt]
   \normalfont \small $^1$FAIR, AI @ Meta &\normalfont \small $^2$Georgia Tech &\normalfont \small $^3$Hello Robot Inc \\
   \normalfont \small $^4$Carnegie Mellon & \normalfont \small $^5$Bielefeld University & \normalfont \small $^6$Ukrainian Catholic University \\
   \normalfont \small $^7$Bosch Center for AI & \normalfont \small $^8$IIITA & \normalfont \small $^9$IIT ISM Dhanbad \\
   \normalfont \small $^{10}$IIITM & \normalfont \small $^{11}$CNAEIT & \\
   \end{tabular}\\
   {\texttt{homerobot-info@googlegroups.com}} 
}

\begin{document}

\maketitle

\begin{abstract}

In order to develop robots that can effectively serve as versatile and capable home assistants, it is crucial for them to reliably perceive and interact with a wide variety of objects across diverse environments. To this end, we proposed \textit{Open Vocabulary Mobile Manipulation} as a key benchmark task for robotics: finding \textit{any} object in a novel environment and placing it on \textbf{any} receptacle surface within that environment. We organized a NeurIPS 2023 competition featuring both simulation and real-world components to evaluate solutions to this task. Our baselines on the most challenging version of this task, using real perception in simulation, achieved only an 0.8\% success rate; by the end of the competition, the best participants achieved an 10.8\% success rate, a 13x improvement.  We observed that the most successful teams employed a variety of methods, yet two common threads emerged among the best solutions: enhancing error detection and recovery, and improving the integration of perception with decision-making processes. In this paper, we detail the results and methodologies used, both in simulation and real-world settings. We discuss the lessons learned and their implications for future research. Additionally, we compare performance in real and simulated environments, emphasizing the necessity for robust generalization to novel settings.

\end{abstract}


\section{Introduction}

The future of in-home assistive robots will require robots that are capable of generalization and reasoning over increasingly complex environments, while manipulating a wide variety of objects, many of which are not well represented in training data. To encourage work in this direction, we proposed the HomeRobot OVMM benchmark~\cite{yenamandra2023homerobot}, which posits that one fundamental problem in robotics and embodied AI is \textit{open vocabulary pick and place:}
 \begin{center}
Move \textit{(object)} from \textit{(start location)} to \textit{(goal location)}
\end{center}
Solving this challenging problem in \textit{any} previously-unseen environment requires bringing together research in perception, planning, and policy learning. Increasingly, we've seen a number of works which solve parts of this problem~\cite{yenamandra2023homerobot,chang2023goat,liu2024okrobot,hu2023toward}, being able to perform multi-step manipulation in previously-unseen environments.

In addition, there have been a number of robotics challenges in the past~\cite{duckietown,robocup,real-robot-challenge}, while a separate, parallel community has looked at simulation-based challenges, such as the Habitat Rearrangement Challenge~\cite{habitatrearrangechallenge2022} and iGibson~\cite{li2021igibson}. Simulation has the advantage of being extremely accessible; regardless of resources, teams can use simulations like Mujoco~\cite{todorov2012mujoco}, ManiSkill~\cite{mu2021maniskill}, or AI2 Thor~\cite{kolve2017ai2}. This makes simulation-based challenges quite appealing, but unfortunately it's not so clear how well simulations transfer to the real world~\cite{restrospectives}; in fact, there's strong evidence they do \textit{not} transfer particularly well~\cite{gervet2022navigating}. At the same time, a proliferation of capable, low-cost robot hardware~\cite{kemp2022design,fu2024mobile} means that real-world experiments are accessible and increasingly reproducible, even if not all labs are using the same platforms yet. 

\begin{figure}[bt]
    \includegraphics[width=\linewidth]{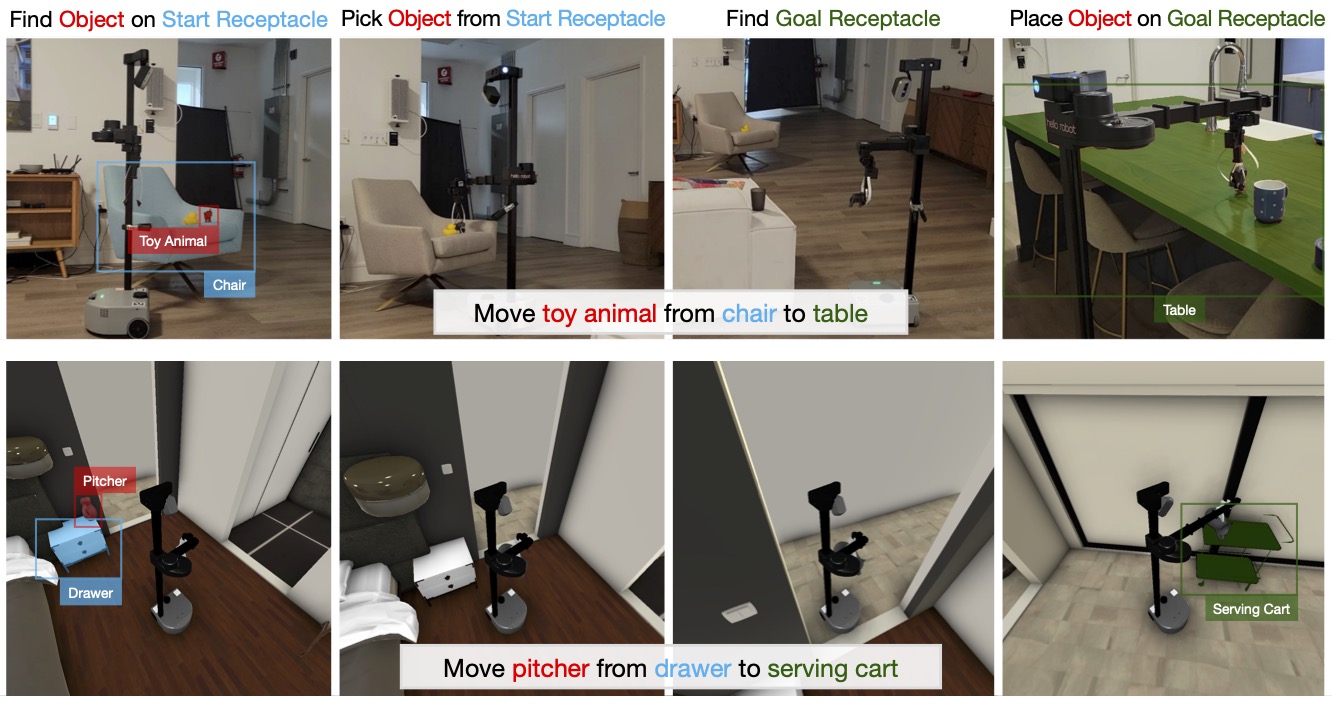} 
    \caption{We have designed simulated homes with articulated and moveable objects. Policies can be trained and evaluated in simulation (bottom row) with the same control stack as on physical hardware (top row) while performing the OVMM task. Agents that can accurately navigate around, search for, and manipulate objects in simulation were transferred to the real world benchmark. Figure from~\cite{yenamandra2023homerobot}.}
    \label{fig:sim-and-real}
    \vspace{-5pt}
\end{figure}

As such, we proposed a competition
with both simulation and physical robot evaluation for open-vocabulary mobile manipulation\footnote{Found at \url{https://aihabitat.org/challenge/2023_homerobot_ovmm/}}.
The aim is to facilitate research that leverages recent advances in machine learning, computer vision, natural language, and robotics to build agents able to navigate in a novel environment and manipulate previously-unseen objects. This is in contrast to most prior work, which assumed a closed world with known classes~\cite{szot2021habitat}.
The real and simulation versions of the challenge are shown in Fig.~\ref{fig:sim-and-real}.
We constructed training and test splits for our task both within the simulation and the real world, using the affordable but capable Hello Robot Stretch platform~\cite{stretch-design,stretch-grasp}. 
To our knowledge, no such competition or benchmark has rigorously demonstrated all of these abilities, but many have tackled facets of the problem \cite{dasari2022rb2,deitke2020robothor,shridhar2020alfred,szot2021habitat}. 


In addition, we provided a modular control stack where advances in perception, grasping, navigation, and so forth, can be tested without requiring the re-implementation of the entire robot control infrastructure~\cite{paxton2023homerobot}. We took an algorithm-agnostic approach when implementing this baseline, providing both reinforcement learning and heuristic baselines related to prior work~\cite{yenamandra2023homerobot}.

The competition spanned $6$ months, with $61$ teams and $79$ submissions, from which three were selected for real-world evaluation. On the technical side, we see that open-vocabulary perception methods were still the main bottleneck for real-world embodied AI, with a very large number of methods failing in challenging real and simulated scenarios~\cite{zhou2022detecting,mobile_sam,fu2024mobile}


Some core takeaways are:
\begin{itemize}[leftmargin=15pt]
    \item \textbf{Current perception models are essential, but are not good enough on their own.} All our participant teams spent time engineering around current open-vocabulary vision models and attempting to improve performance.
    \item \textbf{Error detection and recovery is crucial.} Especially since perception is not reliable from larger distances, it's very important that agents be constantly planning and reasoning over what they can try next.
    \item \textbf{Use Docker to share code and reproduce results.} Docker-based evaluations allow us to try the same methods on other robots, and to try the same methods in simulation vs. the real world. Especially as robotic systems grow more complicated, it's difficult to easily reproduce code without containerization.
    \item \textbf{How best to use robot learning in open environments is still not clear.} It did not seem like reinforcement learning resulted in less engineering effort; in fact, tuning RL reward functions was quite difficult, and resulted in a lot of issues for our teams, vs. running a classical robotics stack.
\end{itemize}

\section{Competition Setup}

The competition is split into two components: the Simulation Challenge and the Real World Challenge.
We first describe the Simulation Challenge in Sec.~\ref{ssec:simulation-challenge}, which serves as the qualifier for inclusion in the Real-World Challenge. The top three entrants were then evaluated in a real environment, as described in Sec.~\ref{sec:real-world}.
Participating entries were evaluated -- both in simulation for the automatic leaderboard and on the physical robot -- using three metrics. Metrics were averaged over evaluation episodes, though winners were chosen based on overall success.

\textbf{Overall Success.} A trial is successful if, at the end of the trial, the specified object is anywhere on top of a target receptacle of the correct category. Anywhere on top of the surface is acceptable.
If at any point the real robot collides with scene geometry, the task fails immediately.

\textbf{Partial Success.}
In addition to the overall success, we report success for each of the four individual sub-tasks: (1) finding the target object on a start receptacle, (2) grasping the object, (3) finding the goal receptacle, and (4) placing the object on the goal receptacle (full success).

\textbf{Steps.} The number of actions taken to solve each episode, a proxy for task completion efficiency.

\begin{figure}[bt]
  \centering
  \begin{subfigure}[b]{0.3\textwidth}
    \includegraphics[width=\textwidth]{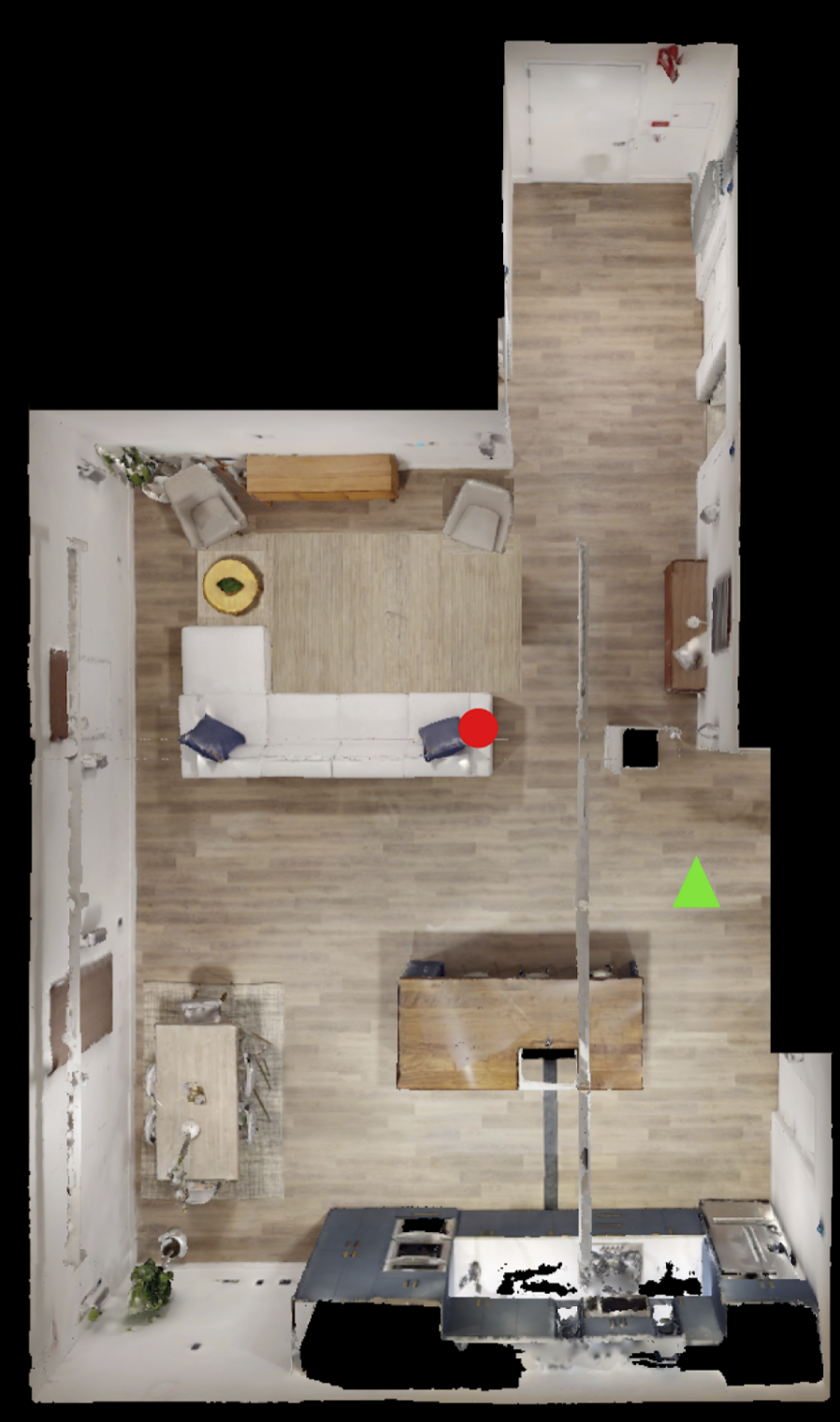}
    \caption{\textit{cup} from \textit{couch} to \textit{chair}}
    \label{fig:map1}
  \end{subfigure}
  \begin{subfigure}[b]{0.3\textwidth}
    \includegraphics[width=\textwidth]{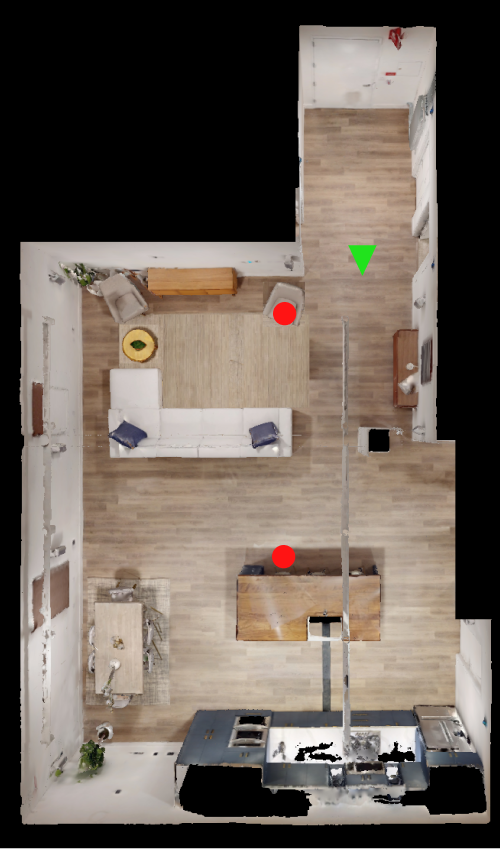}
    \caption{\textit{toy animal} from \textit{chair} to \textit{table}}
    \label{fig:map2}
  \end{subfigure}
  \begin{subfigure}[b]{0.3\textwidth}
    \includegraphics[width=\textwidth]{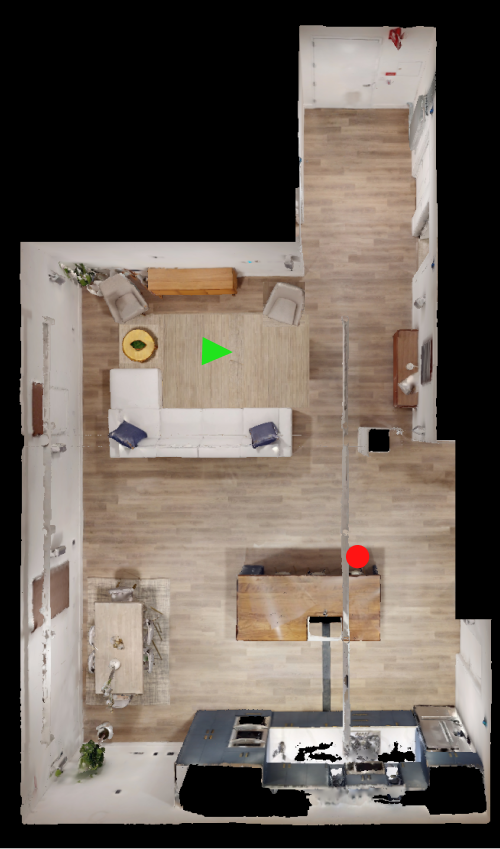}
    \caption{\textit{lemon} from \textit{chair} to \textit{table}}
    \label{fig:map3}
  \end{subfigure}
  \caption{Maps used for real-world testing and evaluation. We ran real-world evaluations on three different environments, testing search, grasping, and placement of different types of objects and given different queries.}
  \label{fig:world-maps}
  \vspace{-5pt}
\end{figure}

\subsection{Simulation Challenge}
\label{ssec:simulation-challenge}

We used Habitat~\cite{savva2019habitat, szot2021habitat} as our simulation platform. Habitat provides a rich environment useful for training control, navigation, perception, and language understanding models. Participants had access to complex cluttered environments populated with diverse objects, receptacles, and an extensive array of clutter assets. The richness of this simulated environment facilitated rapid iteration and training cycles, specially enabling inexpensive data generation for reinforcement learning-based policies. 

We used EvalAI\footnote{\url{https://eval.ai/}} to host the challenge, a popular platform for evaluating machine learning problems which has hosted over 200 AI challenges with over 180k submissions. Participants uploaded Docker images containing their agents, as described in Sec.~\ref{sec:docker}, which were then 
evaluated on an AWS GPU-enabled instance.
EvalAI provided a public leaderboard that will show all participant submissions.

For the challenge, we created an episode dataset consisting of 5 splits: train, val, minival, test standard and test challenge, using a total of 60 scenes from the Habitat Synthetic Scenes Dataset (HSSD)~\cite{khanna2023habitat}. We released the train, val, and minival episode splits, which were created from the train and val scenes consisting of 38 and 10 scenes, respectively. These splits allowed the participants to train and evaluate their agents locally. The minival and test splits were hosted on EvalAI. The test splits contained previously unseen object categories and scenes which were held out from public training and validation datasets, in order to ensure that our participants' methods were actually generalizing to unseen objects and environments. Each of the hosted split (referred to as phase by EvalAI) had its own leaderboard and served a distinct purpose:

\begin{enumerate}[leftmargin=15pt]
    \item \textit{Minival} split: The purpose of this split was to confirm that remote evaluation ran properly and reported the same result as local evaluation. Each team was allowed up to 100 submissions per day, on a very small set of validation episodes.
    \item \textit{Test Standard} split: The purpose of this split is to serve as the public leaderboard establishing the state of the art. This is what should be used to report results in papers. Each team is allowed up to 10 submissions per day. 
    \item \textit{Test Challenge} split: This split was used to decide the teams that would move to the next split of the challenge. Each team was allowed a total of 5 submissions until the end of the challenge submission split. Results on this split were not be made public until the announcement of final results at NeurIPS 2023.
\end{enumerate}

Agents were evaluated on 250 episodes and had a total available time of 48 hours to finish each run. All evaluations used an AWS EC2 p2.xlarge instance which has a Tesla K80 GPU (12 GB Memory), 4 CPU cores, and 61 GB RAM. The use of Docker containers for evaluation, a hidden test set, and the constraint on the number of official challenge submissions helped to prevent overfitting on the test set while allowing broad participation. 

\begin{figure}[bt]
  \centering
  \begin{subfigure}[b]{0.2\textwidth}
    \includegraphics[width=\textwidth]{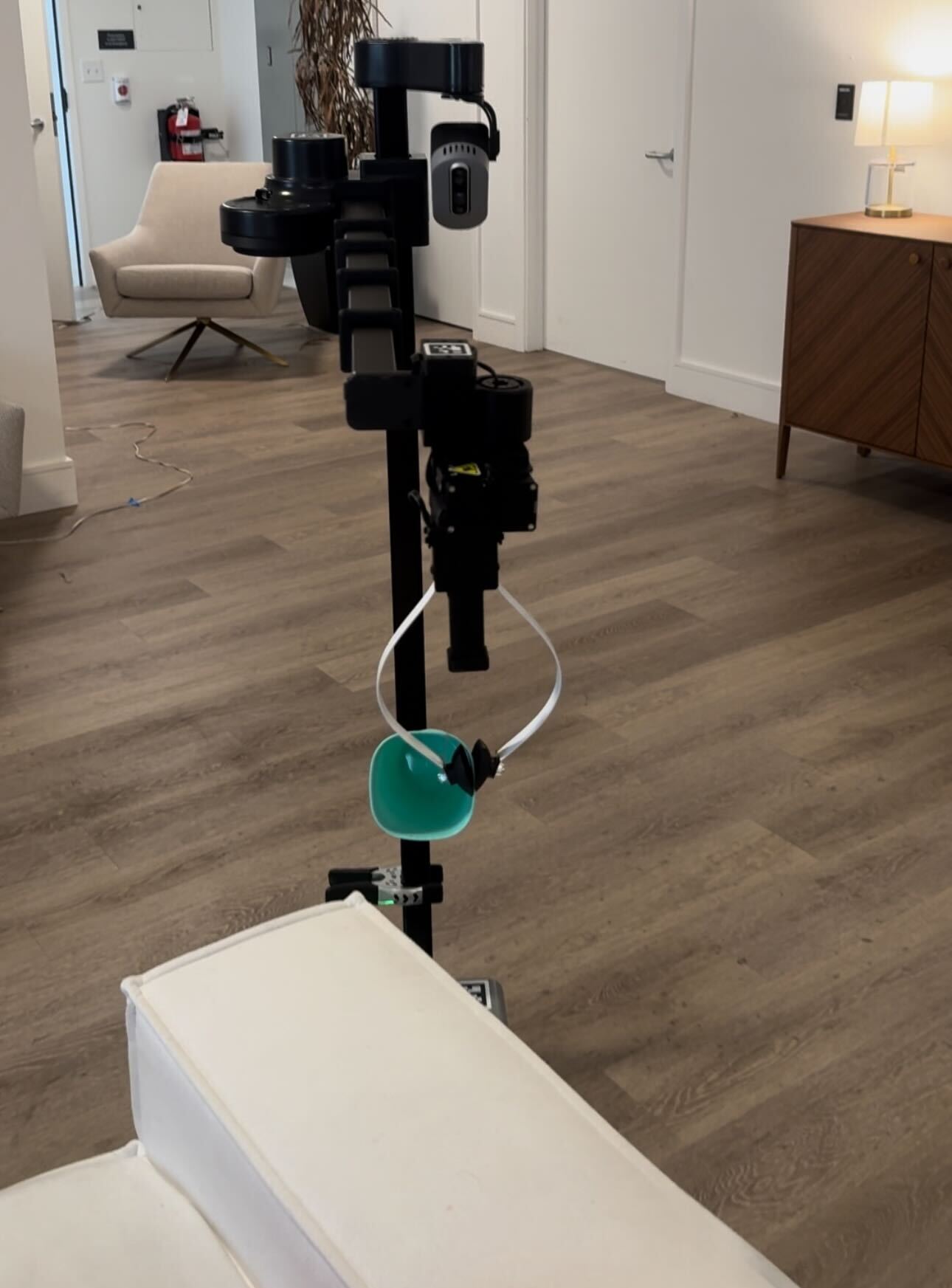}
     \caption{
     \textit{cup} from \textit{couch} to \textit{chair}}
    \label{fig:real1}
  \end{subfigure}
  \begin{subfigure}[b]{0.2\textwidth}
    \includegraphics[width=\textwidth]{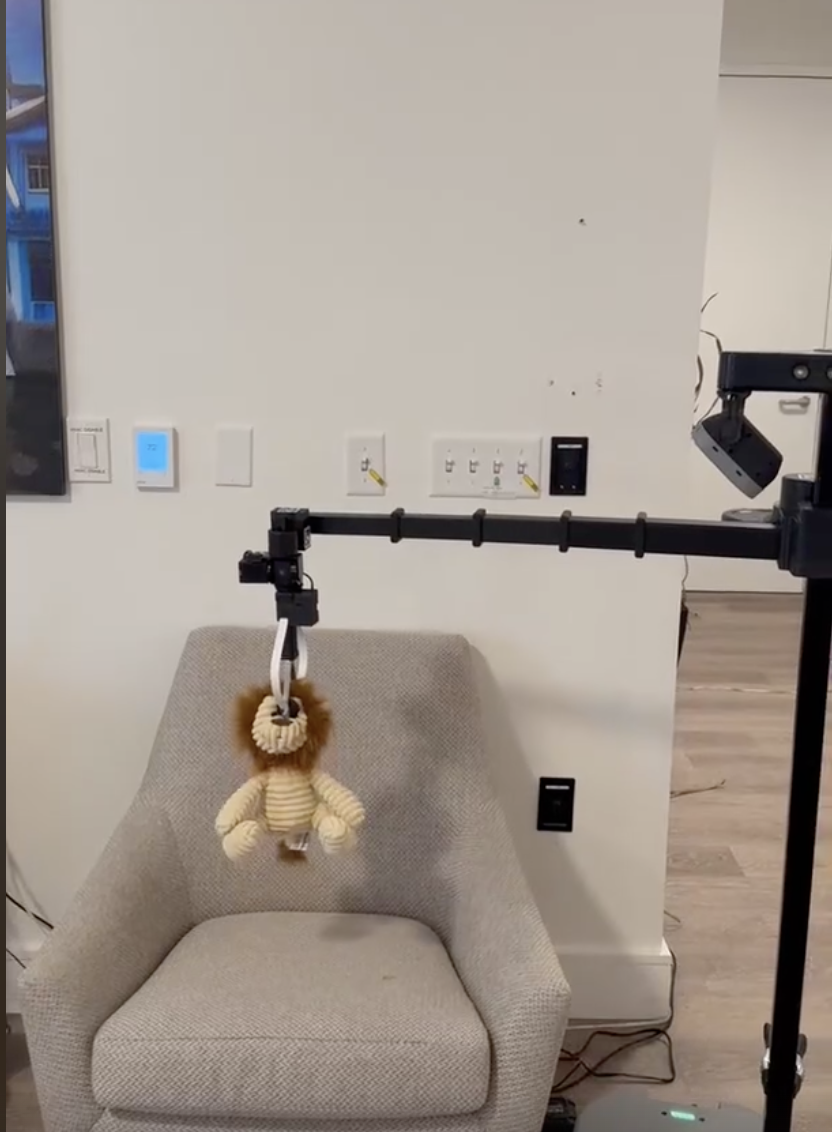}
     \caption{
     \textit{toy animal} from \textit{chair} to \textit{table}}
    \label{fig:real2}
  \end{subfigure}
  \begin{subfigure}[b]{0.2\textwidth}
    \includegraphics[width=\textwidth]{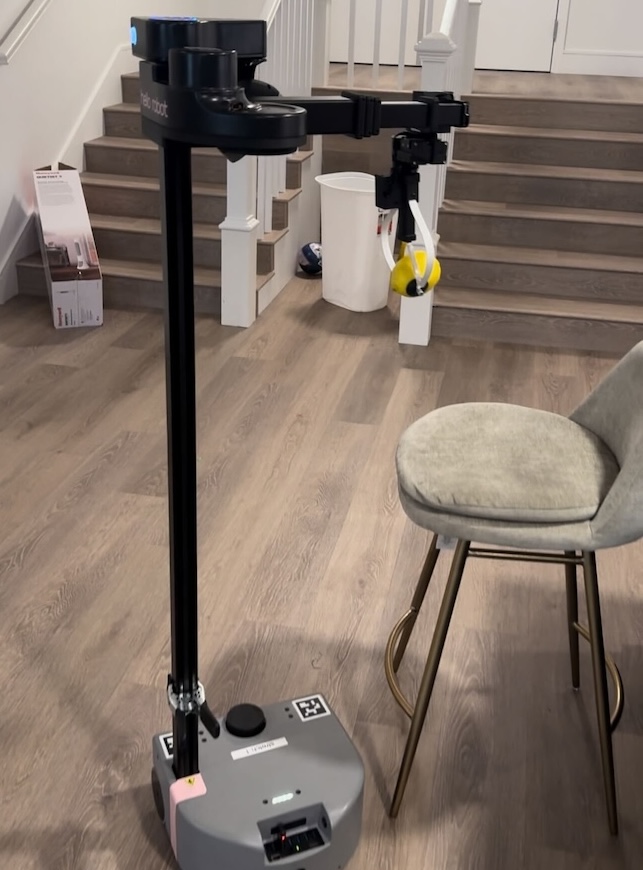}
     \caption{
     \textit{lemon} from \textit{chair} to \textit{table}}
    \label{fig:real3}
  \end{subfigure}
  \begin{subfigure}[b]{0.35\textwidth}
    \centering \includegraphics[width=\textwidth]{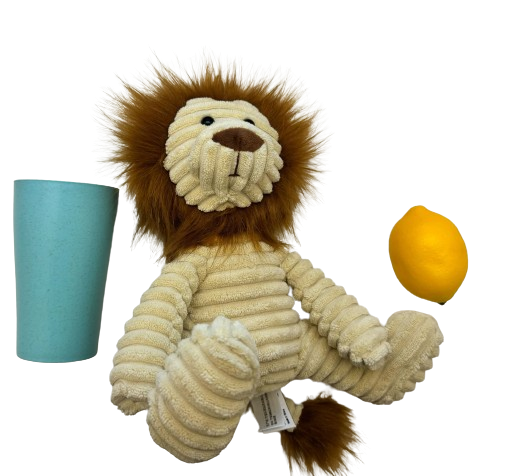}
     \caption{Goal objects used for real-world evaluations.}
    \label{fig:allobjects}
  \end{subfigure}
  \caption{Examples of object manipulation from real-world experiments. The participant teams needed to find and move three objects (\textit{fig~\ref{fig:allobjects}}) from the held-out apartment environment: a plastic cup, a toy animal, and a yellow toy lemon.}
  \label{fig:realworld}
  \vspace{-5pt}
\end{figure}

\subsection{Real Robot Challenge}
\label{sec:real-world}

At the end of the simulation challenge phase, the top three teams from the test-challenge leaderboard were selected for the real robot challenge. We used Hello Robot's Stretch RE2 robot~\cite{kemp2022design} to evaluate the agents in the real environment. The stretch 2 is a lightweight, low-cost mobile manipulator which can fit into a variety of different environments and is broadly suitable for homes. 
 Additionally, as noted, participants do not need access to robot hardware to participate in the challenge.

We ran three evaluations per team in a controlled real-world apartment. The layouts for these different environments were shown in Fig.~\ref{fig:world-maps} and snapshots from the experiments are shown in Fig.~\ref{fig:realworld}. We ran a total of $\sim$10 hours of experimentation and evaluations to measure our real-world metrics and determine the results of the experiments. Docker containers were used to leverage using the same environment and code from simulations for a seamless transition. 

\subsection{Containerized Testing}
\label{sec:docker}

\begin{figure}[bt]
    \includegraphics[width=\linewidth]{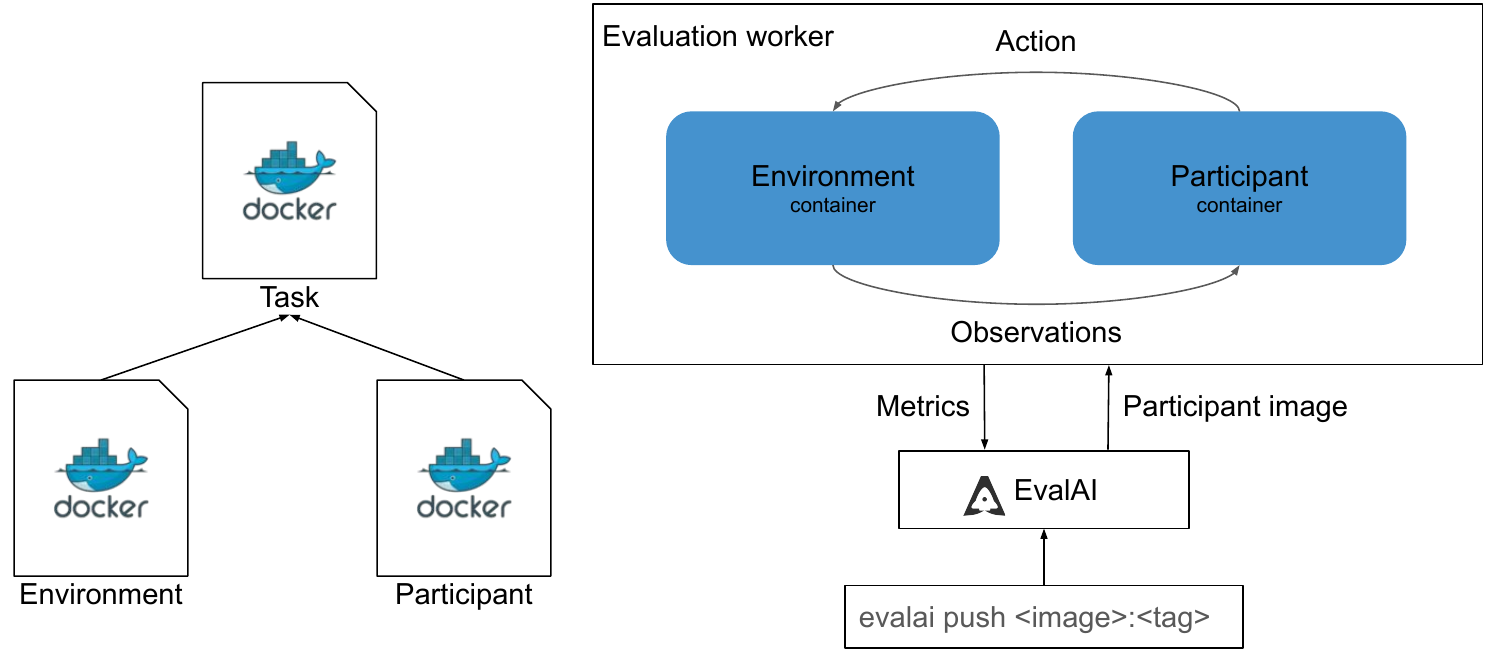} 
    \caption{Left: challenge Docker files hierarchy. Both the participant and environment docker images are built off of the same image; the environment image actually contains the task itself, and the participant docker image sends actions and receives observations over the network. Right: challenge evaluation infrastructure.}
    \label{fig:evalai-pipeline}
    \vspace{-5pt}
\end{figure}
In order to support running the same experiments on the real hardware as in simulation, and reduce setup times/bugs, we had users submit their agent in a Docker image. We set up the challenge evaluation pipeline using three Docker images (see Fig.~\ref{fig:evalai-pipeline}):
\begin{itemize}
    \item \texttt{Task} image: This served as a base Docker image, pre-installed with all libraries required by the software stack. This image includes a gRPC-based interface enabling communication between the participant Docker container and the environment Docker container. Participants are required to use the \texttt{task} image as the base for their submissions.
    \item \texttt{Environment} image: This is a private image which builds upon the \texttt{task} image by replacing public dataset paths with private dataset paths and initiating the environment service, a gRPC wrapper on top of the simulation environment.
    \item \texttt{Participant} image: This image also uses the \texttt{task} image as its base, adding the participant's specific approach implementation, which have to be encapsulated in the agent interface (with act and reset methods). Additionally, we provide an example \texttt{participant} image that launches home-robot baselines.
\end{itemize}

When the participating teams upload their solution Docker images, EvalAI stores them in the AWS Docker registry and schedules an evaluation of their submissions. Then, the platform spawns an evaluation worker that runs the \texttt{environment} and \texttt{participant} Docker containers. The two containers exchange sensor observations and actions using gRPC until the evaluation is over. Finally, the evaluation results are collected and displayed on the leaderbord (see Fig.~\ref{fig:evalai-pipeline}).

To evaluate the teams in the real-world setup, we were able to pull submitted \texttt{participant} Docker images from AWS for the respective teams and run the code.



\subsection{HomeRobot Library}

In order to provide participants a starting point, we provided an open-source library titled HomeRobot~\cite{yenamandra2023homerobot,paxton2023homerobot} containing implementations for two baseline systems: a heuristic system based on prior work~\cite{gervet2022navigating} and a reinforcement learning (RL) baseline~\cite{yenamandra2023homerobot}.

As a part of the modular system, we provided code for: (1) simple grasping, (2) inverse kinematics, (3) frontier-based exploration~\cite{yamauchi1997frontier}, (4) RL policy evaluation, (5) open-vocab detection using Detic~\cite{zhou2022detecting}, (6) simple robot-base motion planning to goals using Fast Marching Method~\cite{sethian1999fast}, and (7) simple whole-body motion planning for manipulation. We provide modular components so that contest entrants can choose as much of the problem as they want to address. For example, if they do not have access to a real robot platform, they could choose to use the discrete grasp planning pipeline implemented by us\footnote{Software stack: \href{https://github.com/facebookresearch/home-robot}{https://github.com/facebookresearch/home-robot}}. 

For training the model-free RL agent, we use the implementation of DD-PPO~\cite{wijmans2019dd} provided by Habitat~\cite{szot2021habitat}. The RL agent uses a different policies for the: (1) navigate to object, (2) gaze, (3) navigate to receptacle and (4) place task. Each policy takes in the output from a sensor and an open-vocabulary object detector as input and output actions for either navigation or rearranging objects.

\section{Results}
In this section, we present the results obtained by the participants in the simulation and real world phases of the OVMM challenge. 

\subsection{Simulation}
The first stage of the OVMM competition was conducted in simulation. The simulation phase started accepting submission on July 12$^{th}$ 2023 and concluded on October 20$^{th}$ 2023. Overall, the simulation phase saw a participation from 61 teams with a total of 368 submission. Out of this, 79 submissions were recorded on the final test-challenge leaderboard by 21 teams. Only $7$ teams surpassed the baselines from our original HomeRobot~\cite{yenamandra2023homerobot} paper. Table~\ref{tab:sim_results} reports the simulation metrics for those teams along with the baseline on the test-challenge leaderboard. The top-performing team~\cite{melnik2023uniteam} achieved an overall success rate of $10.8\%$ significantly outperforming next best team by $+7.6\%$  overall success. In Table~\ref{tab:test_standard_data} we report the results on the test-standard split. 

\begin{table}[h]
    \centering
    \begin{adjustbox}{width=\textwidth,center}
    \begin{tabular}{@{}l c c c@{}}
    \toprule
    \textbf{Team Name} & \textbf{Overall Success} & \textbf{Partial Success} & \textbf{Avg. steps}\\
    \midrule
        UniTeam~\cite{melnik2023uniteam}                & 10.8 & 32.8 & \phantom{0}993.2 \\
        Rulai                                           & \phantom{0}3.2 & 16.8 & 1044.3 \\
        KuzHum (RL YOLO+DETIC (full tracker + )         & \phantom{0}2.4 & 22.2 & 1139.3 \\
        LosKensingtons (Longer)                         & \phantom{0}2.0 & 17.0 & 1075.6 \\
        GXU-LIPE-Li (xtli312/maniskill2023:heuristi)    & \phantom{0}1.6 & 13.9 & 1035.4 \\
        GVL                                             & \phantom{0}1.6 & 12.5 & 1109.0 \\
        VCEI                                            & \phantom{0}1.2 & 15.0	& 1,058.7 \\
        \midrule
        Baseline (Heuristic) & \phantom{0}0.8 & 17.7 & \phantom{0}962.2 \\
        Baseline (RL) & \phantom{0}0.0 & 12.7 & 1203.6 \\
    \bottomrule
    \vspace{-0.2em}
    \end{tabular}
    \end{adjustbox}
    \caption{Simulation results on test-challenge set for the top 6 teams, as well as the baselines from our original HomeRobot paper~\cite{yenamandra2023homerobot}. We saw a dramatic improvement over our baseline methods over the course of the competition, although the best team, UniTeam~\cite{melnik2023uniteam}, still achieved only a fairly low success rate due to the difficulty of the problem.}
    \label{tab:sim_results}
    \vspace{-5pt}
\end{table}

\subsection{Real World}
During evaluations, we picked out the objects and placed them in locations marked as a red dot in Fig \ref{fig:world-maps} and the robot in the position and orientation as the green arrow. Once the setup configuration was set, we ran the docker image containing the multi-step policy code to explore the environment, navigate to goal receptacle, pick, navigate to place receptacle, and place. When exploring, the robot would occasionally collide with several furniture objects, so we set a limit of \textit{3} collisions per evaluation. If this number was reached, the trial would be halted, and the episode would get partial success based on the step it reached.

\begin{table}[h!]
    \centering
    \begin{adjustbox}{width=\textwidth,center}
    \begin{tabular}{l c c }
    \toprule
    \textbf{Team Name} & \textbf{Overall Success} & \textbf{Partial Success}\\
    \midrule
        UniTeam~\cite{melnik2023uniteam} & 33.3\% & 100.0\%  \\
        Rulai & 0.00\% & 66.6\% \\
        KuzHum (RL YOLO+DETIC (full tracker + ) & 0.00\% & 33.3\%  \\
        \midrule
        Baseline (Heuristic) & 33.3\% & 63.3\%  \\
        Baseline (RL) & 66.6\% & 86.7\%  \\
    \bottomrule
    \vspace{-0.2em}
    \end{tabular}
    \end{adjustbox}
    \caption{Real-world results for the top 3 teams. We assigned partial success if the episode was halted or a step failed. The simulation results were predictive of the real-world results as UniTeam~\cite{melnik2023uniteam} was able to get a successful run.}
    \label{tab:real_results}
    \vspace{-5pt}
\end{table}

\section{Analysis and Discussion}

We saw a large diversity of different types of solutions attempted, from purely motion planning based to reinforcement learning to imitation learning. We will describe the results and methodology of the top three performing teams~\cite{melnik2023uniteam,kuzma2024homerobot}, and then describe overall trends. We also performed a survey of the participant teams, in order to capture thoughts going beyond this.
More detailed summaries of the different teams' performance are in Appendix Sec.~\ref{sec:uniteam}, Sec.~\ref{sec:rulai}, and Sec.~\ref{sec:kuzhum}. We additionally ran a survey of our participants.

\paragraph{Perception}

Perception was one of the biggest issues that our teams saw. They used a variety of solutions: Detic~\cite{zhou2022detecting}, as in the original paper, YOLOv8~\cite{yolov8_ultralytics, limberg2022yolo}, OwL-ViT~\cite{minderer2023scaling}, and variants of Segment Anything~\cite{kirillov2023segment} including MobileSAM~\cite{mobile_sam} and LangSam\footnote{\url{https://github.com/luca-medeiros/lang-segment-anything}}. 
Some methods made use of foundation models like CLIP~\cite{radford2021learning}, while others did not at all and just used off-the-shelf detection and motion planning. On their own, these models were not sufficient; many of our teams note detecting previously-unseen objects as the most challenging part.

Taking confidence into account was not enough. UniTeam noted that often, \textit{goal receptacles} in particular could be predicted with very high confidence, even if they were incorrect.

\paragraph{Task Model}

\begin{figure}
  \centering
  \includegraphics[width=1.0\textwidth]{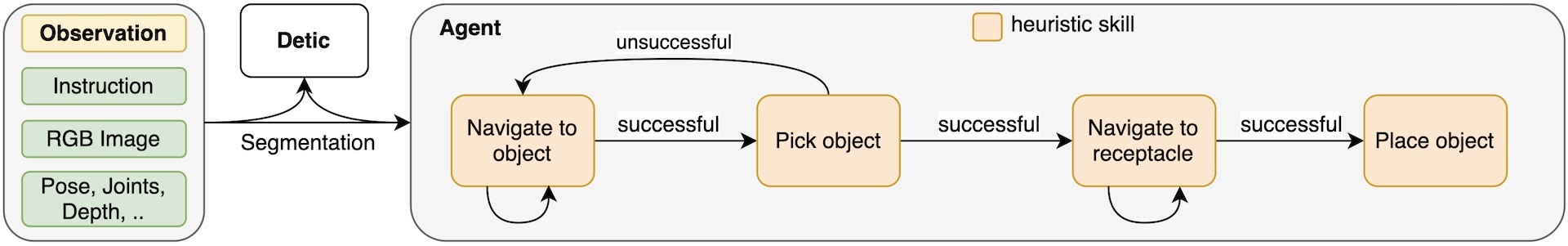}
  \caption{\label{fig:uniteam-architecture} The architecture of \textit{UniTeam}'s agent. It is build as a state machine with four states: \textit{(Find and) Navigate to object}, \textit{Pick object}, \textit{(Find and) Navigate to end receptacle}, and \textit{Place object}. In the \textit{Find}-phases, the \textit{Exploration} skill is used. The agent's perception model is Detic \cite{zhou2022detecting}, it segments the object and receptacles in the RGB image. This segmentation, together with other observations such as the pose, joints and a depth image, are the input to the agent.}
  \vspace{-5pt}
\end{figure}

One key aspect of many of the solutions we see was properly representing the task. Our baselines used a very simple ``task model,'' wherein we would attempt to execute each of four skills in sequence with no retrying: find an example of the target object on the start receptacle, pick it up, find a goal receptacle, place. Many of our contest entrants correctly noted this was a major problem.
For example, see Fig.~\ref{fig:uniteam-architecture}, used by our winning team~\cite{melnik2023uniteam}. They used the same object detection system that we did in our baselines, but they add many steps where the robot determine if it has been successful, and will either try again or revert to an earlier step in the task plan. Team Rulai similarly made retrying past skills a priority (see Appendix~\ref{sec:rulai}). 

Determining when and how to retry failed skills is an important area for future research; there is evidence that for example vision-language models could play a strong role in this going forward, versus manual engineering, for automatic construction of behavior with error recovery~\cite{zhang2023grounding,ahn2024vader}.

\paragraph{Manipulation}
Manipulation skills were a challenge for all of the teams. Reward engineering for improving the placement skill was very difficult, as was predicting and avoiding collisions between the arm and the environment with motion planning. Teams spent time improving placement However, the basic skills were able to transfer to the real world reasonably well.

\paragraph{Simulation vs Real World}
None of the teams who responded to the survey had access to a real-world Stretch robot, which included all of the top three teams chosen for real-world testing, which suggests that the simulation is useful for testing and development.

\paragraph{The Role of Learning in Successful Solutions}
Notably, the winning team did not use any learning at all, though two of the top three teams did some amount of finetuning. Teams commmented that they did not have time to make learning solutions work, compared to improving model-based solutions. Future competitions should probably allocate more time.

\section{Running the Challenge: Lessons Learned}

\paragraph{Enhancing Communication for Future Challenges.} 
For future challenges, we recommend enabling better, faster, and more reliable communication channels. One survey response indicated a lack of information about challenge changes, such as deadlines or the number of submissions: “we lacked information about challenge changes, like a deadline or the number of submissions.” Another participant expressed frustration with late-breaking changes: “the most frustrating thing about the OVMM competition was deadline postponement in the last 24 hours of the challeng\dots as by the time of it we nearly spent all of the submission attempts.”

Other challenges have successfully used tools like Discord\footnote{\url{https://discord.com/}} to build responsive and active communities, allow users to assist one another, and facilitate rapid communication. Discord supports quick interactions through threads and direct messages and has been employed by research projects~\cite{liu2024okrobot}, companies and open-source initiatives. Unfortunately, we did not adopt this approach, but we recognize its potential. For future contests, we recommend using a platform like Discord instead of a mailing list to improve communication with participants.

\paragraph{Distributed Evaluation.} 
One limitation of our approach was the reliance on a single evaluation environment in the real world. Although this was dictated by the high cost of setup, it resulted in a bottleneck for scaling real-world evaluation, constrained by the availability of the robot and the evaluation apartment. To address this, a potential recommendation for future challenges is to implement distributed evaluation systems, which would enhance both diversity and resilience.

\paragraph{Containerized Testing.}
The use of Docker for submissions in both simulation and real-world testing proved highly effective. This approach allowed us to test identical policies in both environments across various countries and labs without encountering dependency issues. This was a significant success, and we strongly recommend continuing with this type of evaluation in the future. It facilitates more real-world and sim-to-real competitions, enhancing the robustness and applicability of the results.

\section{Conclusions}

We consider the competition to have been a success, with a mixture of interesting lessons. There is substantial work to be done in improving perception, error recovery, and in making sure robot learning scales to real-world environments. We also identified some promising tools: simulations do allow people to write useful robot code, which can be tested in previously-unseen environments, which makes robotics research more accessible to everyone. In addition, tools like the containerized testing setup we deployed make reproduction across teams and labs much easier.

\anon{
\section{Acknowledgement}
\textbf{UniTeam} would like to thank  the Ministry of Culture and Science of the State of North Rhine-Westphalia, Germany, for their support through the KI-Starter funding program, which allowed us to test our approaches in this challenge.
}

\bibliographystyle{plainnat}
\bibliography{references}
\newpage

\appendix

\section{Team UniTeam}
\label{sec:uniteam}

\anon{\textbf{Members:} Andrew Melnik, Michael Büttner, Leon Harz, Lyon Brown, Gora Chand Nandi, Arjun PS, Gaurav Kumar Yadav, Rahul Kala, and Robert Haschke.}

UniTeam's method is mostly based on the heuristic baseline of the OVMM challenge with additional improvements~\cite{melnik2023uniteam}. They used Detic~\cite{zhou2022detecting} for detection, and made a number of specific improvements to navigation, exploration, and manipulation skills. They (in this section "we") largely did not use any learning in their solution.

\subsection{Overview}

The  agent's capabilities are divided into the following key skills:
\begin{enumerate}
    \item \textbf{Detection}: Object detection and segmentation on the agent's camera RGB-D images. Building \textit{Bird's Eye View} (BEV) map with semantic areas.
    \item \textbf{Exploration}: Exploring the environment to find the start location (\texttt{start\_receptacle}) and the goal location (\texttt{end\_receptacle}). Exploring \texttt{start\_receptacles} to find the object.
    \item \textbf{Navigation}: Move towards the specified navigation goal: the object, start- or end receptacle.
    \item \textbf{Picking}: Picking up the object, supported by a high-level action command due to the absence of simulated gripper interaction in Habitat.
    \item \textbf{Placing}: Adjusting the agent's position in front of the \texttt{end\_receptacle} and placing the object.
\end{enumerate}

The skills are executed in a certain order. At the beginning, the agent turns 360° to get an overview of its surroundings. After that, the agent uses the Exploration skill to find the object, navigates towards it if the object is found and uses the Picking skill to pick it up. Should the Picking procedure be unsuccessful, the agent goes back to use the Exploration skill to find the object. After the object was successfully picked, the agent uses either the Exploration skill again if no entity of the \textit{end receptacle} was found, or it skips directly to the Navigation skill to move to the nearest \textit{end receptacle}, and ends with the Placing skill to place the object. 

The Detection skill is executed in every frame of the run. This means that the agent collects information of \textit{start} and \textit{end receptacles} even if the current goal is to find the object. The baseline uses the perception model Detic~\cite{zhou2022detecting} to generate masks for objects and receptacles.
An overview of our \textit{UniTeam}-agent's architecture can be found in Figure~\ref{fig:uniteam-architecture}.

\subsection{Improvements over the Baseline}

UniTeam decided to iteratively analyze the problems of the baseline and find solutions for them. 

\subsubsection{Detection}
Initially, the baseline agent employed the same confidence threshold value for all receptacles and objects. During testing, it became evident that the agent frequently overlooked objects with this threshold value. However, reducing the threshold value increased the misclassification rate of \textit{end receptacles}. The implementation of a dynamic object/receptacle-specific threshold proved effective in mitigating this issue. 
This allowed for a lower confidence threshold for objects and a higher confidence threshold for \textit{end receptacles}, enabling the detection module to identify objects more reliably and reducing misclassifications of \textit{end receptacles}. 

\begin{figure}
  \centering
  \includegraphics[width=0.5\textwidth]{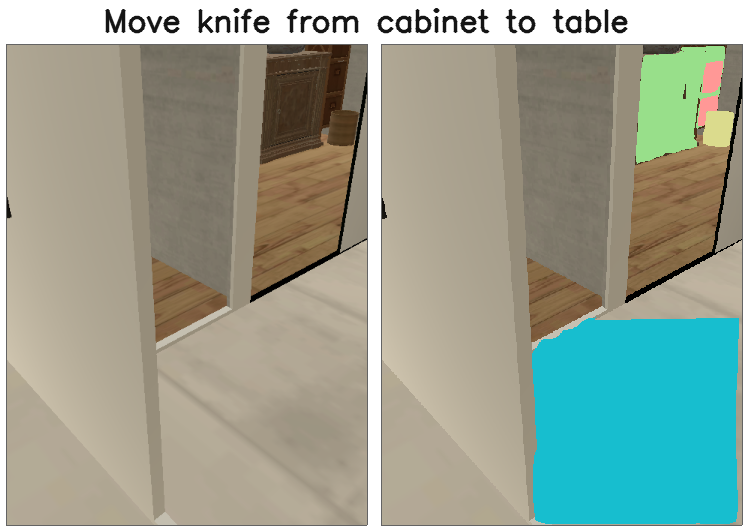}
  \caption{\label{floor} Example of an incorrect floor classifications. The floor is classified as a table (blue cluster), which is the \textit{end receptacle}. The floor is filtered out in the \textit{UniTeam}-agent, otherwise the agent would want to place the knife on there.}
\end{figure}

The perception module also encounters challenges due to incorrect detection of
objects on the floor or the floor itself (see Fig.~\ref{floor}) as objects or receptacles. To minimize the occurrence of these false positives \cite{melnik2021critic}, we established a height threshold for receptacles within the perception module using the depth information. This eliminates detections corresponding to the floor level, thereby facilitating more accurate planning.

\subsubsection{Exploration}
The baseline agent frequently encountered challenges when strictly adhering to a single goal, as the heuristic consistently selected the same action, leading to infinite oscillations between two adjacent states. To mitigate this issue, a decision-making mechanism was introduced to operate on the goal map. This mechanism allows the agent to choose between the goal point and exploring the frontier during navigation, preventing persistent commitment to an unattainable goal \cite{melnik2018world}.

In the baseline approach, the agent directs itself towards the nearest identified receptacle. However, due to imperfect object detection, the nearest receptacle may not align with the intended goal. To address this, our agent only stores the most probable \textit{end receptacles} in the Bird's Eye View (BEV) map \cite{harter2020solving}, depending on their confidence score. Subsequently, the agent selects the reachable receptacle with the highest matching probability, reducing the likelihood of placing the object on incorrectly classified receptacles. As the agent begins adding identified \textit{end receptacles} to the goal map during the \textit{Navigate to object} phase, a sufficient number of \textit{end receptacles} are typically available for selection.

\subsubsection{Navigation}
 When the baseline agent detected a potential collision, the collision was not incorporated efficiently into the new planning, leading the agent to attempt the same movement as before. To rectify this behavior, the detected collision is now integrated at the last short-term goal on the obstacle map. This refinement provides a more detailed representation of the environment in the BEV map, enabling the agent to formulate improved plans. Due to the strategy followed by the baseline agent, instances arose where the agent was not well-aligned with the \textit{end receptacle}, resulting in an inadequate position for successful object placement. To address this limitation, the navigation in our agent aims to position the agent near the center of the goal receptacles. This adjustment ensures generally improved starting positions for the object placement routine.


\subsubsection{Picking}

The baseline picking routine necessitates the presence of the target object within the field of view of the agent's camera. However, in certain scenarios, the object is not visible anymore after the agent moved towards the receptacle, thus resulting in picking failures. To address this issue, our agent rotates left and right if the initial picking attempt fails, expanding its overall field of view of the receptacle and thereby increasing the probability of object detection.

In our agent a check has been introduced to verify whether the object was successfully picked up. Only upon a successful pick, the agent transitions to the next phase; in case of a failed pick, however, our agent reverts to the navigation phase, exploring different locations for the object.

\subsubsection{Placing}
The baseline agent encountered challenges in accurately calculating the distance between the agent and the placing point, resulting in attempts to move either too far away or too close to the receptacle. This issue was mitigated in our agent by moving the agent toward the receptacle in small steps over multiple iterations, ensuring that the agent is positioned appropriately for successful object placement.

Additionally, the baseline agent often placed objects close to the edges of receptacles, causing the object to fall down. To address this, the placing point was adjusted to maintain a safe distance from the receptacle edges, preventing later object falling down.

In situations where the detection module does not recognize the \textit{end receptacle}, the baseline agent previously placed the object blindly, resulting in failures due to collisions or misplaced objects. In such cases, our agent places the object on any flat surface in front of him, assuming that they belong to the \textit{end receptacle}. This reduces the occurrence of misplacements.

Furthermore, in the placing phase, large objects grasped by the baseline agent frequently collide with the \textit{end receptacle}. This issue is alleviated in our agent by dropping the object onto the \textit{end receptacle} from an increased height.

\subsection{Panorama Update}

After the challenge concluded, UniTeam worked on a new approach which utilizes panorama images.
As discussed above, the baseline and their submitted architecture used Detic~\cite{zhou2022detecting} to do segmentation on the RGB image of each timestep. This means that in some cases important objects or receptacles could be missed because the camera didn't happen to point towards them during exploration, or because they are located at the edge of the image and could not be classified correctly. These problems can be solved by using panorama images. The idea is that by creating panorama images regularly, the agent gets a good overview over large parts of the scene without missing important areas.

To create a panoramic image, we rotate the robot by 360° while taking images, do an equirectangular transformation of each image and stitch these images together. To minimize stitching artifacts, we turn the robot’s camera by 180° before creating a panorama image, ensuring the camera lens remains closely aligned with the rotation axis.
We used OWL-ViT v2~\cite{minderer2023scaling} as a detection module to create bounding boxes and MobileSAM\cite{mobile_sam} to segment the insides of these bounding boxes.

A new approach is introduced to decide when to create a panorama image. The agent starts each episode with a panorama image. While updating the map, the agent saves the places which have been covered by the panorama image, taking a maximum depth limit into account. During exploration, the agent moves towards unexplored parts of the map. When such a part is reached, another panorama image is taken. This ensures that all parts of the apartment are covered in a panorama image. Since the agent would often take panorama images at door frames, which provide an unoptimal angle for the subsequent room, we let the agent move into the unexplored area a bit before taking another panorama image.

\subsection{Major Issues}

The most prominent issue UniTeam ran into was the correct detection of objects and receptacles. In most episodes that failed, the issue was that either the object or the receptacles were not detected, or that the wrong objects and receptacles were incorrectly classified as the correct one. The latter issue is still a problem even with our approach of saving the confidence scores, since incorrect \textit{end receptacles} could still have a high confidence. The precision of the detection model varies depending on the specific object or receptacle.

Another major issue we ran into was arm collisions while placing and picking. While attempting either of the procedures, the robot would extend the arm and sometimes hit either miscellaneous objects on the receptacles or the receptacle itself. Other issue we frequently encountered were not having the object in the camera frame once during the episode and thus not finding the object, \textit{end receptacles} being too large to place objects on them without collisions, or the robot getting stuck at some part of the environment. 
We also encountered issues with the simulation itself. For example there was a scene with a sink in a bathroom that was not marked as a sink in the ground truth. Thus, attempts to place objects on them would fail. Another issue arose when the robot places objects like pens on the correct \textit{end receptacles}. The objects would roll back and forth on the receptacles forever without coming to rest, which is a requirement for counting as a success.

\section{Team Rulai}

\anon{\textbf{Members:} Yang Luo, Jinxin Zhu, Yansen Han, Bingyi Lu, Xuan Gu, Qinyuan Liu, Yaping Zhao, Qiting Ye, Chenxiao Dou, and Yansong Chua.}

Team Rulai mostly based their method on the heuristic baseline.
In the OVMM competition, they divided the vision-language task into three parts: detection, navigation, and manipulation (picking and placing). Our method is based on domain adaptive semantic segmentation and policy-based mobile manipulation
\begin{itemize}
	\item{\textbf{Detection}}: To enhance the performance in indoor scenes, we build up a mixed dataset by combining simulator images with real images in common datasets like LVIS\cite{DBLP:conf/cvpr/GuptaDG19}, and select VLDet\cite{DBLP:conf/iclr/LinSJ0QHY023} as the baseline detection model to finetune.
	\item{\textbf{Navigation}}: The pipeline is built upon the heuristic baseline. We propose strategies for autonomous escape, target region selection, and accelerated exploration.
	\item{\textbf{Manipulation}}: To approach a reachable location with enough space to execute the placement action, the agent explores open areas and finds an optimal placement point.
\end{itemize}

\subsection{Improvements over the Baseline}
The team made improvements to detection, navigation, and manipulation in order to achieve their second-place performance.

\subsubsection{Detection} 
Rulai used two approaches to improve detection performance:

1) Mixed data finetuning: The baseline model is trained on large open-source images rather than indoor scenes of the OVMM task, thus we collect indoor images from common datasets and mix them with images observed from the simulation platform to finetune our baseline model. 

2) Baseline model selection: The OVMM task has a requirement of running time, therefore a suitable baseline model needs to balance performance and response time. We re-implement eight vision-language detection models and compare them within a benchmark constructed from images we extracted from Habitat-Sim, and finally choose the one that performs best within the time of competition requirement.

\subsubsection{Navigation}
Team Rulai made three major changes to navigation to improve performance over the baseline:
\begin{enumerate}
    \item They implemented a self-rescue strategy to prevent the agent from getting stuck in one place and wasting steps. 
    \item They designed a rapid exploration algorithm based on prior state encoding to reduce the overall number of operations. 
    \item They added an algorithm to identify the optimal target navigation area in order to satisfy the conditions for convergence.
\end{enumerate}

\subsubsection{Manipulation}
After the navigation to the receiver task is completed, the agent will stop at a fixed distance from the placement container. This position may not necessarily be the best place to perform placement actions, therefore, we first need to adjust it to reach the optimal executable placement position. We rotate the robot 360 degrees with a fixed number of steps and determine if the agent can see the appropriate placement area in the field of view after each rotation. If a suitable placement area is not found after the rotation ends, switch back to navigation to another receiving container.

After seeing the appropriate placement area in the field of view, the agent should adjust the orientation of the host towards the target area and move to the nearest position to stop.

After reaching the closest position to the target area, in order to ensure the success rate of placement, it is necessary to explore the optimal placement point. We match the mask area of the target area with the footprint of the placement object to find the appropriate and optimal placement point.

\subsection{Major Issues}

Team Rulai noted a few issues with their solution.

\textbf{1) Detection:} 
One key issue is the word-object mismatch due to the non-custom baseline model and the low-fidelity simulator images. They observed that the detection model trained on a general dataset performs poorly in the simulated environment, resulting in a high false positive rate. Mixed data training may alleviate this problem while preserving perception ability in the real world. Methods to further fill the reality gap are worth exploring.


\textbf{2) Navigation:}
 Unknown environment: The agent cannot know the layout and positions of obstacles in advance, which increases the difficulty of path planning and navigation. 2) No distance sensors: The lack of distance sensors means the agent cannot directly measure the distance to obstacles or the target, limiting its ability to perceive the environment. 3) Imprecise target navigation area: The inaccuracy of the target location makes it challenging for the agent to determine an exact navigation route, increasing the difficulty of finding the target.

\textbf{3) Manipulation:}
After the navigation task is completed, the position where the agent stops may not be the best place to perform placement actions. If the position of the agent host is not adjusted, the agent may fail to place due to the inability of the robotic arm to reach the target placement position. At the end of the navigation task, the agent does not determine the best placement area, which can lead to objects being placed in inappropriate positions, causing them to slide or shake. After finding a suitable placement area, it is also necessary to determine the optimal placement point within the area. It is necessary to place the object in the middle of the target container or in an open area to further avoid the risk of slipping.
\label{sec:rulai}

\section{Team KuzHum}
\label{sec:kuzhum}

\anon{\textbf{Members:} Volodymyr Kuzma, Vladyslav Humennyy, and Ruslan Partsey.}

Team KuzHum's agent was based on the RL baseline, provided by the organizers. We introduce an improved perception module that uses fine-tuned YOLOv8 \cite{yolov8_ultralytics} model for detection of known object classes, combined with segmentation by MobileSAM \cite{mobile_sam}, and Detic \cite{zhou2022detecting} detection and segmentation model for unseen classes. We enable object tracking in the YOLO model to achieve more robust segmentation, similar to ground truth that was used for training of the policies. We also utilize fine-tuned place module, trained with modified reward function, and a small change in global skill sequence that repeats navigation to object and pick object skills if the object was not picked after a try.

\subsection{Improvements over the Baseline}

KuzHum's improvements focused on detection, as well as improving the reinforcement learning skills, and allowing for retrying of various high-level skills.

\subsubsection{Detection}

During initial analysis of the baseline, we observed a significant impact of accurate semantic segmentation on the success of the baseline agent: switch from ground truth segmentation to Detic was followed by a big drop in success rate of all skills. As expected, the perception module of the agent influenced every skill and overall episode completion, therefore, during our work we mainly focused on its improvement.

\begin{figure}[bt]
    \centering
  \includegraphics[width=0.9\linewidth]{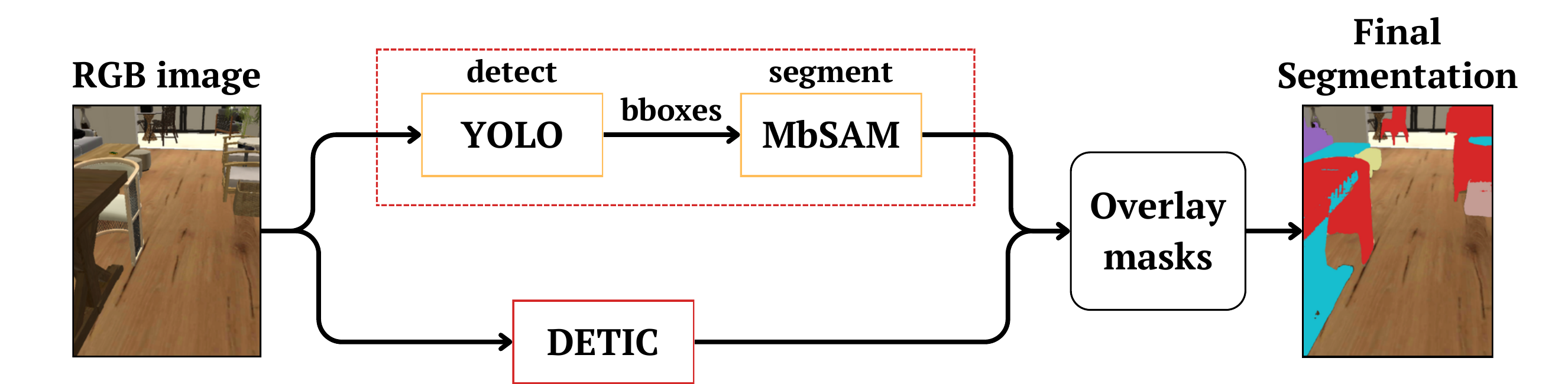}
    \caption{Semantic segmentation pipeline used by Team KuzHum. They overlaid masks from two different models in order to improve performance.}
    \label{fig:kuzhum-segmentation}
\end{figure}

We enhanced the baseline segmentation module, which initially utilized only Detic, by adding an extra pipeline using the MobileSAM model for segmentation. This model is prompted with bounding boxes from YOLOv8 (as shown in Fig. \ref{fig:kuzhum-segmentation}). The masks from both modules are combined, with task-specific objects overlaid on top of other detected objects. This combination allows us to preserve the open-vocabulary Detic segmentations while incorporating task-specific YOLO-SAM, which is trained to distinguish furniture types and small objects.

\subsubsection{RL Checkpoint Finetuning}

Besides improving the segmentation models, we also focused on fine-tuning the model's baseline RL checkpoints. For the Place phase, we customized the training reward to encourage the agent to keep the object in the camera view and move closer to the goal receptacle. These additional changes were motivated by the low success rate of the place skill, which directly impacted the overall success of the task.

\subsubsection{Task Model}

During the evaluation, we observed that the agent prematurely ended its exploration after performing an unsuccessful Gaze skill. This led to abruptly ended episodes, even if the agent had not failed. However, since we had a sensor indicating whether the object was successfully picked, we could implement conditional behavior for the agent based on success or failure. This prompted us to add a few improvements:

\begin{enumerate} 
    \item Multiple trials for the Gaze skill: If the agent initiates the Gaze skill, it attempts it up to 10 times unless it succeeds. 
    \item If the agent fails the Gaze skill 10 times, it reverts to the Navigate to object skill. 
\end{enumerate}

That change to heuristic led to more robust algorithm, which could handle faulty detections of goal object and consequent gaze skill call.

\subsection{Major Issues}

The most problematic and challenging task was fine-tuning of RL based place policy. Initially, the policy reward was sparse, giving almost all positive feedback to the agent after the actions were finished: for successful drop and stable position of an object on the receptacle. This led to inefficient iterations during training, requiring much more resources and time for improvements than we could find. Subsequent modification of the reward, described in the previous part, had a positive impact at first but caused the success rate to drop after we further prolonged training. The agent got stuck in the local maximum of the reward function, exploiting it while not achieving successful placement (it sometimes dropped the object right from the start of the episode while, on average, achieving a higher reward than ever before). Debugging of the reward terms did not yield any positive results, so we had to drop most of the training results and use the weights with moderate improvement over baseline. Also, some problems in the agent's behavior that were addressed by modifications of reward, such as frequent blocking of the camera view by the manipulator, remained.

\section{Simulation Results on Test-Standard Split}
In Table~\ref{tab:test_standard_data}, we provide the results of the evaluation of the Docker for all the teams on the test-standard split. Comparing it to the test-challenge set (Table~\ref{tab:real_results}), we see that UniTeam is still the best performing team, while Rulai had attained the 4$^{th}$ position on test-standard. Only 6 of the teams were able to get more than $1.0\%$ on the leaderboard which shows the difficulty of the challenge.

\begin{table}[h]
    \centering
    \begin{adjustbox}{width=\textwidth,center}
        \begin{tabular}{@{}clccc@{}}
            \toprule
            \textbf{Rank} & \textbf{Team Name}        & \textbf{Overall Success}  & \textbf{Partial Success}  & \textbf{Avg. Steps}\\
            \midrule
            1    & UniTeam          & 6.0            & 27.6            & 1048.8  \\
            2    & KuzHum (RL)      & 2.8            & 19.1            & 1153.0  \\
            3    & PieSquare        & 2.0            & 11.1            & 1082.6  \\
            4    & Rulai            & 1.6            & 11.1            & 1037.3  \\
            5    & cucumber         & 1.2            & 11.2            & \phantom{0}943.5   \\
            6    & VCEI             & 1.2            & 12.3            & 1058.5  \\
            7    & PoorStandard (rl\_deterministic\_self\_trained) & 0.8 & \phantom{0}9.3 & 1218.1 \\
            8    & Clear            & 0.4            & 13.4            & \phantom{0}961.3   \\
            9    & scale\_robotics  & 0.4            & 10.7            & 1061.2  \\
            10   & extreme ProArt (v0) & 0.4         & 11.3            & 1070.0  \\
            11   & GXU-LIPE-Li      & 0.4            & 10.6            & 1063.2  \\
            12   & USTCAIGroup (baseline\_sam\_rl\_with\_heuristic) & 0.4 & 10.9 & 1211.1 \\
            13   & KuDA             & 0.4            & 11.5            & 1219.2  \\
            14   & HomeRobot Challenge Organizers (rl) & 0.4 & 10.9 & 1196.8 \\
            15   & T-1511           & 0.0            & 10.5            & 1063.7 \\
            16   & LosKensingtons   & 0.0            & 10.5            & 1063.7  \\
            17   & USC-GVL          & 0.0            & 10.5            & 1063.7  \\
            18   & GVL              & 0.0            & 10.5            & 1063.7  \\
            19   & HomeRobot Challenge Organizers (heuristic) & 0.0 & 10.5 & 1063.7 \\
        \bottomrule
        \end{tabular}
    \end{adjustbox}
    \vspace{0.2em}
    \caption{Simulation Challenge results on the test-standard set for all the teams that successfully submitted their Dockers on EvalAI. UniTeam is still the top team on the test-standard, while Rulai does not even feature in the top 3.}
    \label{tab:test_standard_data}
\end{table}

\section{Full Set of Test Environments}

\begin{figure}[htbp]
  \centering
  \begin{subfigure}[b]{0.23\textwidth}
    \includegraphics[width=\textwidth]{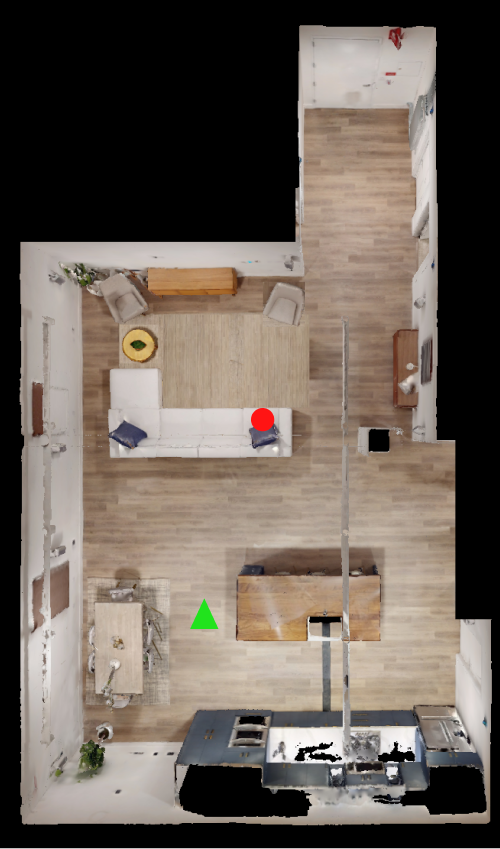}
    \caption{cup couch chair}
    \label{fig:image1}
  \end{subfigure}
  \begin{subfigure}[b]{0.23\textwidth}
    \includegraphics[width=\textwidth]{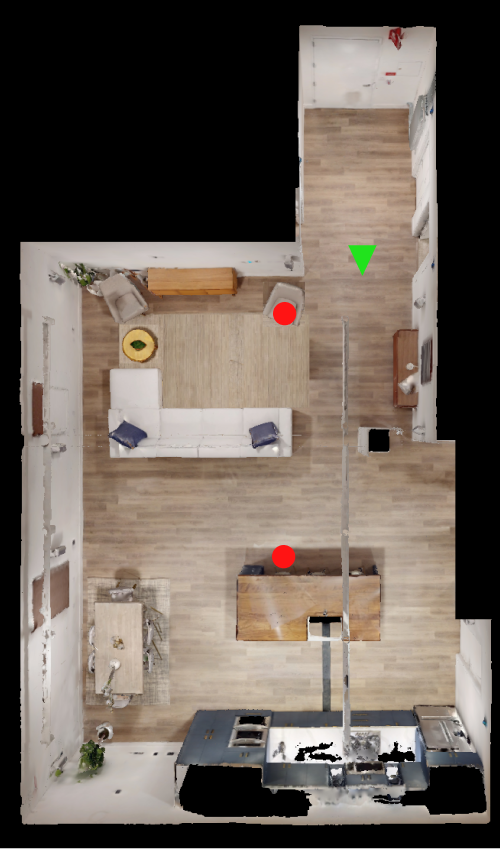}
    \caption{toy animal chair table}
    \label{fig:image2}
  \end{subfigure}
  \begin{subfigure}[b]{0.23\textwidth}
    \includegraphics[width=\textwidth]{images/task_3.png}
    \caption{lemon chair table}
    \label{fig:image3}
  \end{subfigure}
  \begin{subfigure}[b]{0.23\textwidth}
    \includegraphics[width=\textwidth]{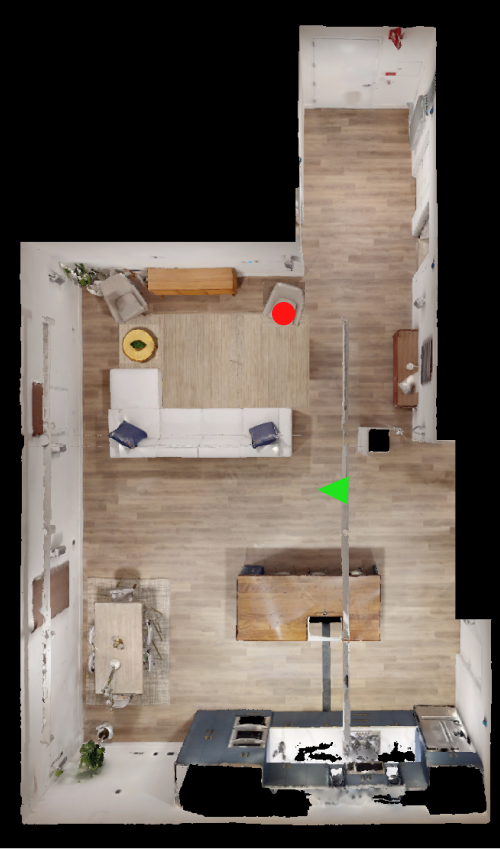}
    \caption{toy animal chair couch}
    \label{fig:image4}
  \end{subfigure}
  \begin{subfigure}[b]{0.23\textwidth}
    \includegraphics[width=\textwidth]{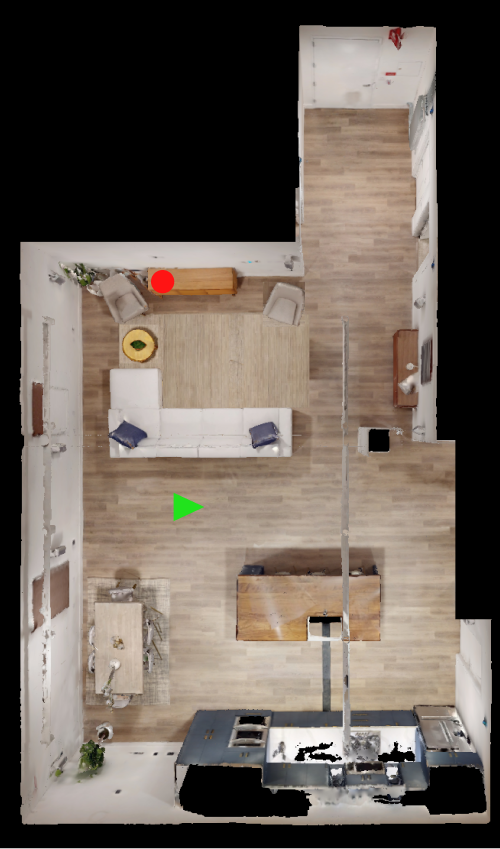}
    \caption{toy animal cabinet couch}
    \label{fig:image5}
  \end{subfigure}
  \begin{subfigure}[b]{0.23\textwidth}
    \includegraphics[width=\textwidth]{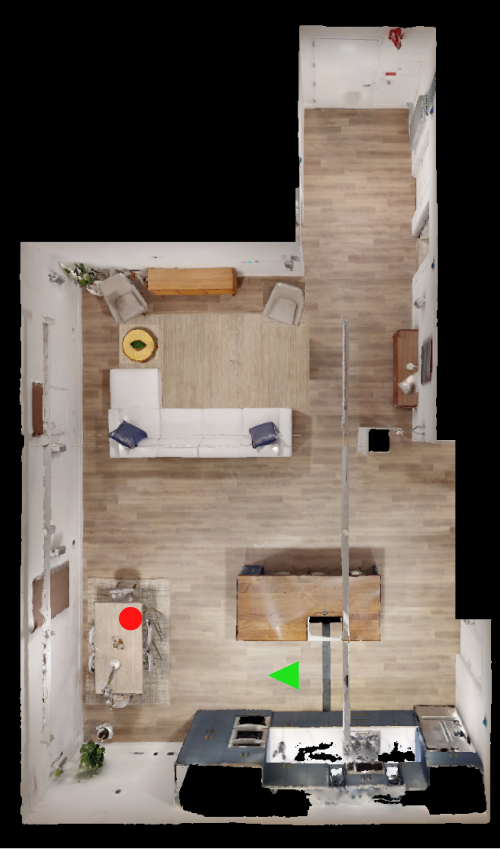}
    \caption{bowl table couch}
    \label{fig:image6}
  \end{subfigure}
  \begin{subfigure}[b]{0.23\textwidth}
    \includegraphics[width=\textwidth]{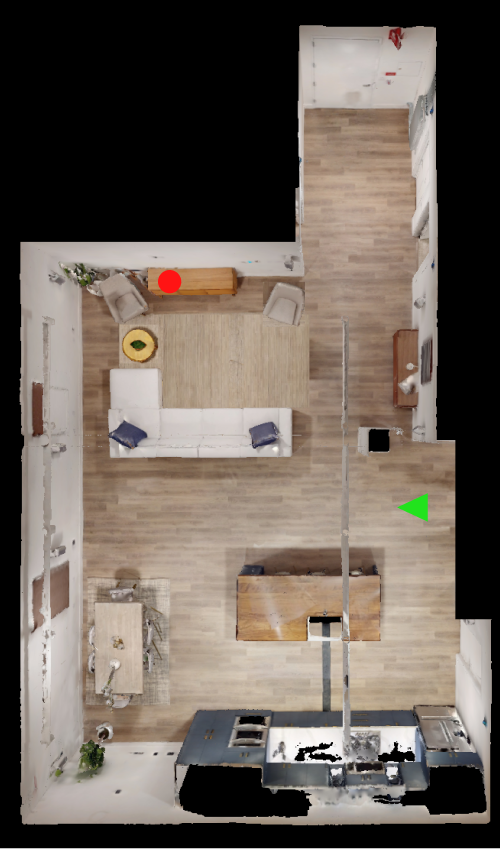}
    \caption{toy drill cabinet table}
    \label{fig:image7}
  \end{subfigure}
  \begin{subfigure}[b]{0.23\textwidth}
    \includegraphics[width=\textwidth]{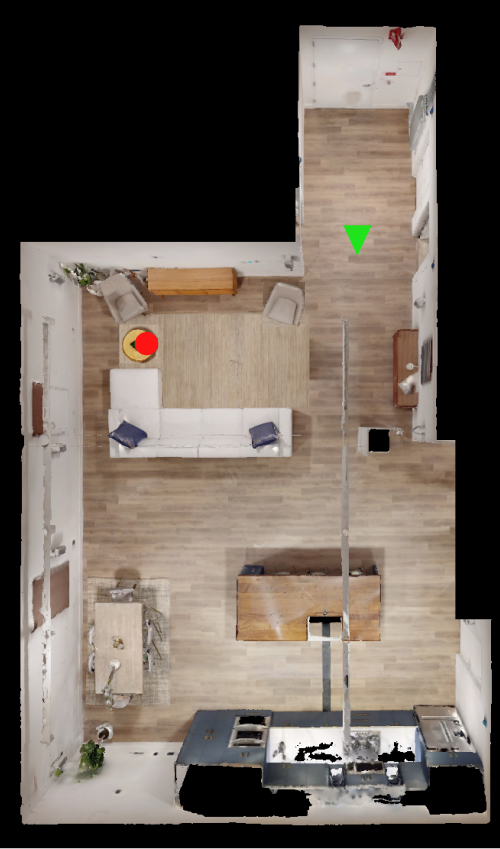}
    \caption{chair lemon table}
    \label{fig:image8}
  \end{subfigure}
  \begin{subfigure}[b]{0.23\textwidth}
    \includegraphics[width=\textwidth]{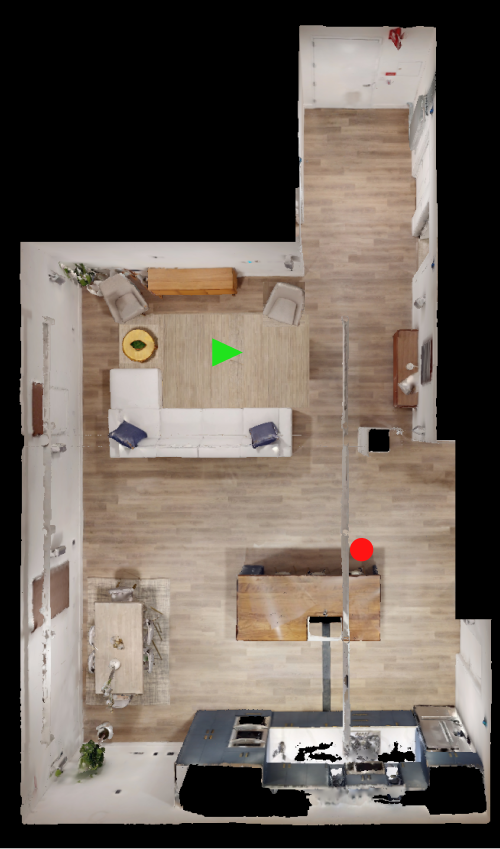}
    \caption{chair lemon table}
    \label{fig:image9}
  \end{subfigure}
  \begin{subfigure}[b]{0.23\textwidth}
    \includegraphics[width=\textwidth]{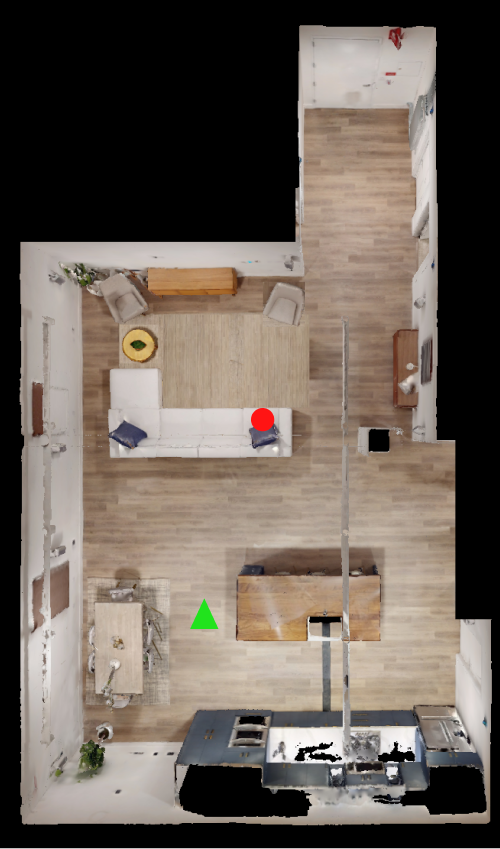}
    \caption{chair lemon table}
    \label{fig:image10}
  \end{subfigure}
  \caption{The ten environments used for real-world testing of the baseline methods, from which we chose three for contest entries. These employ a combination of seen and unseen objects, including ones that did not appear in test or training data.}
  \label{fig:full-world-maps}
\end{figure}

Fig.~\ref{fig:full-world-maps} shows all environments that were used for baseline testing and development during the real-world robot experiments. Due to time constraints, we were only able to test on three of these for competition, given in Fig.~\ref{fig:world-maps}.

\section{Survey Questions}
Our survey floated as a Google Form and sent to challenge participants involved the following questions:
\begin{itemize}
    \item Team Name
    \item Which organizations are your team members affiliated with?
    \item Describe your method in a few lines.
    \item What kind of approach was your method based on?
        \begin{itemize}
            \item Reinforcement learning based
            \item Imitation learning based
            \item Path planning based
            \item Mixed
            \item Other (mention in textbox)
        \end{itemize}
    \item For navigation, did you build off of motion planning or RL?
        \begin{itemize}
            \item Motion Planning
            \item Reinforcement Learning (including our pretrained skills)
            \item Imitation learning
            \item Other (mention in textbox)
        \end{itemize}
    \item For manipulation, did you build off of motion planning, use hand-tuned heuristics, or RL?
        \begin{itemize}
            \item Heuristics (including ours)
            \item Reinforcement learning (including our pretrained skills)
            \item Motion planning
            \item Imitation learning
            \item Other (mention in textbox)
        \end{itemize}
    \item Which foundation models did you use?
        \begin{itemize}
            \item CLIP
            \item VC-1
            \item R3M
            \item GPT3/4/etc
            \item None
            \item Other (mention in textbox)
        \end{itemize}
    \item Did you change the object detection approach?
        \begin{itemize}
            \item Yes
            \item No
        \end{itemize}
    \item Which model did you use to detect open-vocabulary object categories?
    \item Did your team own a Stretch robot and try your method in the real world at the time of the challenge?
        \begin{itemize}
            \item Yes
            \item Did not own Stretch
            \item Did not try in real world
        \end{itemize}
    \item Do you have a report/manuscript associated with your submitted approach? If yes, please provide the link.
    \item What was the biggest surprise when attempting the challenge?
    \item How was your overall experience with the challenge? Rate 1-5
    \item Are you participating in the follow-up challenge?
    \item If no, why not?
        \begin{itemize}
            \item I am participating
            \item No time
            \item No prize
            \item No sim-to-real
            \item OVMM problem was too hard
            \item OVMM problem was not hard enough
        \end{itemize}
    \item Any other feedback on the OVMM challenge?
\end{itemize}

\end{document}